\documentclass[conference]{IEEEtran}
\IEEEoverridecommandlockouts

\usepackage{cite}
\usepackage{array}
\usepackage{booktabs}
\usepackage{amsmath,amssymb,amsfonts}
\usepackage{algorithmic}
\usepackage{graphicx}
\usepackage{textcomp}
\usepackage{xcolor}
\def\BibTeX{{\rm B\kern-.05em{\sc i\kern-.025em b}\kern-.08em
    T\kern-.1667em\lower.7ex\hbox{E}\kern-.125emX}}
\begin{document}

\title{Meta-Learning Based Few-Shot Graph-Level Anomaly Detection\\
}

\author{\IEEEauthorblockN{1\textsuperscript{st} Liting Li}
\IEEEauthorblockA{\textit{School of New Media and Communication} \\
\textit{Tianjin University}\\
Tianjin, China \\
llt19980126@163.com}
\and
\IEEEauthorblockN{2\textsuperscript{nd} Yumeng Wang\textsuperscript{*}}
\IEEEauthorblockA{\textit{School of New Media and Communication} \\
\textit{Tianjin University}\\
Tianjin, China \\
ymwang@tju.edu.cn}
\and
\IEEEauthorblockN{3\textsuperscript{rd} Yueheng Sun}
\IEEEauthorblockA{\textit{College of Intelligence} \\
\textit{Tianjin University}\\
Tianjin, China \\
tud2023@sina.com}
}

\maketitle

\begin{abstract}
Graph-level anomaly detection aims to identify anomalous graphs or subgraphs within graph datasets, playing a vital role in various fields such as fraud detection, review classification, and biochemistry. While Graph Neural Networks (GNNs) have made significant progress in this domain, existing methods rely heavily on large amounts of labeled data, which is often unavailable in real-world scenarios. Additionally, few-shot anomaly detection methods based on GNNs are prone to noise interference, resulting in poor embedding quality and reduced model robustness. To address these challenges, we propose a novel Meta-Learning-based Graph-Level Anomaly Detection framework (MA-GAD), incorporating a graph compression module that reduces the graph size, mitigating noise interference while retaining essential node information. We also leverage meta-learning to extract meta-anomaly information from similar networks, enabling the learning of an initialization model that can rapidly adapt to new tasks with limited samples. This improves the anomaly detection performance on target graphs, and a bias network is used to enhance the distinction between anomalous and normal nodes. Our experimental results, based on four real-world biochemical datasets, demonstrate that MA-GAD outperforms existing state-of-the-art methods in graph-level anomaly detection under few-shot conditions. Experiments on both graph anomaly  and subgraph anomaly detection tasks validate the framework's effectiveness on real-world datasets.

\end{abstract}

\begin{IEEEkeywords}
Graph neural network(GNN), graph anomaly detection(GAD), few-shot learning(FSL).
\end{IEEEkeywords}

\section{Introduction}

With the increasing use of graph data across various fields such as social network analysis, protein structure recognition in bioinformatics, and cybersecurity threat detection, anomaly detection has gained significant attention. Detecting rare and significantly deviating observations is crucial for tasks such as fraud prevention and early detection of cybersecurity threats. 
Graph-level anomaly detection, namely subgraph anomaly detection\cite{qiao2024deep, AS-GAE} and graph anomaly detection\cite{qiao2024deep, GLocalKD}, plays a key role for capturing abnormal behaviors in graph data. 

\textbf{Graph anomaly detection} focuses on identifying anomalous behavior across entire graphs by detecting individual graphs whose structural characteristics significantly deviate from expected patterns. 
Recent research on graph anomaly detection is primarily two main methods: GNN-based approaches ~\cite{dou141, GLAD, qiu2024self} and network representation learning methods \cite{GLocalKD,graph2vec144,FGSD145}. 
GNN-based approaches, such as GLADC \cite{GLAD}, employs contrastive learning to capture graph-level anomalies by reconstructing node features and structures. Network representation learning methods include approaches like GLocalKD \cite{GLocalKD}, combines both global and local anomaly information through knowledge distillation.

\textbf{Subgraph anomaly detection} focuses on identifying local structural anomalies within a graph, where these anomalies often have a broader scope and are typically clustered, making them more challenging to detect compared to individual nodes or edges. 
With the rise of deep learning, Graph Neural Networks (GNNs) have been developed to detect anomalous subgraphs by learning structural and attribute information. For example, DeepFD\cite{DeepFD} focuses on fraud detection by identifying suspicious dense blocks in bipartite attribute graphs. AS-GAE\cite{AS-GAE} uses a position-aware autoencoder to reconstruct graphs, enabling the detection of anomalous subgraphs.

Although subgraph anomaly detection and graph anomaly detection focus on different types of anomalies, they both involve detecting abnormal sets of nodes and edges. 
Existing methods for graph-level anomaly detection face several challenges: \textbf{lack of labeled anomalies}, where limited labeled data and high annotation costs lead to overfitting and reduced generalization; \textbf{noise interference}, where irrelevant graph information affects aggregation and decreases robustness; and \textbf{lack of prior anomaly knowledge}, as similar subgraphs containing valuable anomaly-related information are often overlooked.

To address these challenges, we propose a Meta-Learning-based Graph-Level Anomaly Detection (MA-GAD) framework. First, the graph is compressed into a smaller version to ensure that a GNN trained on the compressed graph performs comparably to one trained on the original graph. A bias loss function is introduced to increase the score gap between anomalous and normal nodes. Additionally, a meta-learning module is incorporated to "learn to learn" from similar networks, capturing relevant anomaly information and training a generalized graph-level anomaly detection model. With minimal data, the model performs well on target graphs.

When applying MA-GAD to anomaly detection, we address both local and global anomalies. Anomalies are detected at both the node and graph levels, improving detection capabilities. The main contributions of this paper are as follows:
\begin{itemize}
    \item We introduce a graph compression module that reduces graph size, preserves essential structural information, and ensures GNN performance on the compressed graph aligns with the original.
    \item We incorporate a meta-learning module that extracts anomaly-related knowledge from similar networks, enhancing model generalization, while a bias loss function increases the separation between anomalous and normal nodes. 
    \item We evaluate the proposed frameworks on graph anomaly detection and subgraph anomaly detection tasks, with experiments on four real-world datasets showing high accuracy and superior performance over traditional baselines in few-shot graph-level anomaly detection.
\end{itemize}

\section{PROPOSED METHOD}
\subsection{Overview}
The MA-GAD framework consists of three steps: (1) A graph compression strategy minimizes performance loss during GNN training. (2) A meta-learning algorithm extracts meta-anomaly knowledge from auxiliary networks for better generalization. (3) A graph anomaly loss function enhances statistical deviation between normal and anomalous nodes, addressing class imbalance from both local and global perspectives.

\subsection{Graph Compression for Performance Preservation}

To reduce the graph size while retaining the effective information, ensuring that the performance of the compressed graph during training is consistent with that of the original graph, this model introduces a graph compression module inspired by the GCOND algorithm~\cite{GCOND}. The goal is to compress the original graph into a smaller synthetic graph and parameterize its structure such that the performance of the GNN model trained on the compressed graph is similar to that trained on the original graph. Gradient matching loss is employed as the compression objective, optimizing the performance of the compressed graph.

Given the graph dataset $G=\left(A,X,Y\right)$, the objective is to learn a compressed graph $\mathcal{K}=\left(A^{\prime},X^{\prime},Y^{\prime}\right)$ that enables similar GNN performance as on $G$. The compressed graph starts with a random distribution $P_{\theta_0}$ and is optimized accordingly:

\begin{equation}\label{equation4_1}
\begin{split}
&\min_{\mathcal{K}}\operatorname{E}_{\theta_0\sim P_{\theta_0}}\left[\mathcal{L}\left(\mathrm{GNN}_{\theta_{\mathcal{K}}}({A},{X}),{Y}\right)\right] \\
&\quad \mathrm{s.t.} \quad \theta_{\mathcal{K}}=\arg\min_{\theta}\mathcal{L}(\mathrm{GNN}_{\theta(\theta_0)}({A}^{\prime},X^{\prime}),Y^{\prime}) ,
\end{split}
\end{equation}
where $\mathcal{L}$ is the cross-entropy loss function, and $\mathrm{GNN}_{\theta}$ represents the GNN model parameterized by $\theta$. This bi-level optimization problem is computationally costly. The gradient matching method from Zhao et al.~\cite{zhao} minimizes gradient differences between compressed and original graphs. A compressed sample set $\mathcal{K}$ is created so that parameters $\theta_{t}^{\mathcal{K}}$ trained on $\mathcal{K}$ at iteration $t$ approximate $\theta_{t}^{G}$ from the original graph. The optimization is:
\begin{equation}
    \label{equation4_2}
    \begin{split}
        &\min_{\mathcal{K}}\mathrm{E}_{\theta_{0}\sim P_{\theta_{0}}}\left[\sum_{t=0}^{T-1}D\left(\theta_{t}^{\mathcal{K}},\theta_{t}^{G}\right)\right] \quad \text{with} \\
        &\theta_{t+1}^{\mathcal{K}} \leftarrow \theta_{t}^{\mathcal{K}} - \eta \nabla_{\theta}\mathcal{L}\left(\mathrm{GNN}_{\theta_{t}^{\mathcal{K}}}(A^{\prime},X^{\prime}),Y^{\prime}\right) \\
        &\quad \text{and} \\
        &\theta_{t+1}^{G} \leftarrow \theta_{t}^{G} - \eta \nabla_{\theta}\mathcal{L}\left(\mathrm{GNN}_{\theta_{t}^{G}}(A,X),Y\right),
    \end{split}
\end{equation}
where $D(\cdot, \cdot)$ is the distance function, and $\eta$ is the learning rate. Gradient matching aligns the training trajectories of the compressed graph with the original, ensuring comparable GNN performance. The optimization is simplified as:
\begin{equation}
    \label{equation4_3}
    \begin{split}
        &\min_{\mathcal{K}}\operatorname{E}_{\theta_0\sim P_{\theta_0}}\left[\sum_{t=0}^{T-1}D\left(\nabla_{\theta}\mathcal{L}\left(\mathrm{GNN}_{\theta_t}(A^{\prime},X^{\prime}),Y^{\prime}\right),\right.\right. \\
        &\quad \left.\left.\nabla_{\theta}\mathcal{L}\left(\mathrm{GNN}_{\theta_t}(A,X),Y\right)\right)\right].
    \end{split}
\end{equation}

To improve gradient matching, the distance function is cosine similarity between gradients of two layers:
\begin{equation}
    \label{equation4_4}
    dis(\mathbf{g}^{\mathcal{K}},\mathbf{g}^{G})=\sum_{i=1}^{d_{2}}\left(1-\frac{\mathbf{g}_{\mathbf{i}}^{\mathcal{K}}\cdot\mathbf{g}_{\mathbf{i}}^{G}}{\left\|\mathbf{g}_{\mathbf{i}}^{\mathcal{K}}\right\|\left\|\mathbf{g}_{\mathbf{i}}^{G}\right\|}\right),
\end{equation}
where $\mathbf{g}_i^{\mathcal{K}}$ and $\mathbf{g}_i^{G}$ represent the $i$-th column of the gradient matrices. As $A^{\prime}$ and $X^{\prime}$ are linked in the graph, the compressed structure can be modeled as a function of compressed node features:
\begin{equation}
    \label{equation4_5}
    \begin{split}
        &A^{\prime} = f_\varphi(X^{\prime}) \\
        &\quad \text{with} \\
        &A^{\prime}_{ij} = \text{Sigmoid}\left(\frac{\text{MLP}_\varphi([X^{\prime}_i;X^{\prime}_j]) + \text{MLP}_\varphi([X^{\prime}_j;X^{\prime}_i])}{2}\right),
    \end{split}
\end{equation}
where $\text{MLP}_{\varphi}$ is a multi-layer perceptron, and $[\cdot; \cdot]$ denotes concatenation. The optimization is then:
\begin{equation}
    \label{equation4_6}
    \begin{split}
        &\min_{X^{\prime},\varphi}\mathrm{E}_{\theta_{0}\sim P_{\theta_{0}}}\left[\sum_{t=0}^{T-1}D\left(\nabla_{\theta}\mathcal{L}\left(\mathrm{GNN}_{\theta_{t}}(f_{\varphi}(X^{\prime}),X^{\prime}),Y^{\prime}\right),\right.\right. \\
        &\quad \left.\left.\nabla_{\theta}\mathcal{L}\left(\mathrm{GNN}_{\theta_{t}}(A,X),Y\right)\right)\right].
    \end{split}
\end{equation}

This requires alternating optimization of $X^{\prime}$ and $\varphi$ due to their interdependence. $\varphi$ is updated for $\tau_1$ iterations, then $X^{\prime}$ for $\tau_2$ iterations, repeating until convergence. In the compressed adjacency matrix $A^{\prime}$, small values below a threshold $\sigma$ are removed to enhance sparsity and efficiency.

After compressing the graph into $\mathcal{K}=(A^{\prime}, X^{\prime}, Y^{\prime})$, a GNN encodes the graph representation directly as: 
\begin{equation}
\label{equation8}
z_G=F_{readout}\left\{f_{GNN}^{2}(A_{l}, f_{GNN}^{1}(A_{l}, X_{l})), v \in V_{l}\right\}.
\end{equation}

\subsection{Loss Function}
A deviation network~\cite{Deviation_loss} builds a loss function to separate normal and anomalous nodes, using anomaly scores ${s_{1},s_{2},\ldots,s_{q}} \sim N(\mu, \sigma^{2})$, where normal scores approximate $\mu=\frac{1}{q}\sum_{i=1}^{q}s_i$ and anomalous scores deviate. The deviation $dev\left(z_{i}\right)=\frac{s_{i}-\mu}{\sigma}$ is integrated into the loss:
\begin{equation} \label{equation4_11} Loss_{i}=(1-y_{i})\cdot|dev(z_{i})|+y_{i}\cdot{max}\left(0,m-dev(v_{i})\right), \end{equation}
where $y_i$ is the label and $m$ the confidence bound.

\subsection{Anomaly Score}
Two fully connected layers form the anomaly evaluation module, processing node and graph embeddings for anomaly scores. The node score is:
\begin{equation} \label{equation4_9} s_{v} = W_{v}^2 \left( \sigma \left( W_{v}^1 z_{v} + b_{v}^1 \right) \right) + b_{v}^2, \end{equation}
where $\sigma$ is ReLU, $W_{v}^1$, $W_{v}^2$ are weights, and $b_{v}^1$, $b_{v}^2$ are biases. The graph score is:
\begin{equation} \label{equation4_9_2} s_{G} = W_{G}^2 \left( \sigma \left( W_{G}^1 z_{G} + b_{G}^1 \right) \right) + b_{G}^2, \end{equation}
where $W_{G}^1$ and $W_{G}^2$ are the weights for the hidden and output layers, and $b_{G}^1$ and $b_{G}^2$ are the bias terms. When performing subgraph anomaly detection, $Loss_i$ is directly used as the loss function. For anomaly graph detection, the loss function is defined as:
\begin{equation} \label{equation4_9_3} Loss = Loss_G + \frac{1}{n} \sum_{i=1}^{n} Loss_i, \end{equation}
where Loss\textsubscript{G} is the graph-level loss. It is formulated using cross-entropy loss to ensure the model accurately distinguishes between normal and anomalous graphs, optimizing the prediction of graph-wide anomaly characteristics.

\subsection{Meta-Learning-based Anomaly Detection}
To boost anomaly detection, the model leverages auxiliary networks’ anomaly knowledge via a meta-learning component inspired by MAML~\cite{MAML}, training initialization parameters $\theta$ for fast adaptation with minimal data. It extracts meta-anomaly knowledge from $k$ auxiliary graphs and fine-tunes on the target graph. The inner loop updates $\theta'_i$ on the support set:

\begin{equation} \theta^{\prime}i = \theta{i} - \alpha \nabla_{\theta_{i}} Loss_{support}^{i}, \end{equation}
where $\alpha$ is the learning rate, and $Loss_{support}^{i}$ is the loss on the support set. The outer loop updates $\theta$ on the query set:
\begin{equation} \theta \leftarrow \theta - \beta \nabla_{\theta_{i}} \sum_{i=1}^{k} Loss_{query} ,\end{equation}
where $\beta$ is the meta-learning rate, and $Loss_{query}$ is the loss on the query set. The pre-trained model is fine-tuned on the target graph for anomaly detection scores.

\subsection{Model Optimization}
To streamline the meta-learning graph anomaly detection module, which involves costly second-order derivatives, two strategies are used. First, the ANIL~\cite{ANIL} algorithm reuses features, updating only the last layer in the inner loop:
\begin{equation}
\label{equation4_15}
\theta_{k}^{(p)}=\left(\theta_{1},\ldots,(\theta_{l})_{k-1}^{(p)}-\alpha\nabla_{(\theta_{l})_{k-1}^{(p)}}\mathcal{L}_{\mathcal{K}_{p}}(f_{\theta_{k-1}^{(p)}})\right).
\end{equation}

Second, reptile~\cite{Reptile} simplifies parameter updates by using $\theta^{0}-\theta^{\prime}$ as the gradient in the outer loop, avoiding second-order derivatives. The inner loop update is:
\begin{equation}
\label{equation4_16}
\theta^{\prime}_i=\theta_{i}-\alpha\nabla_{\theta_{i}}Loss_{support}^{i}.
\end{equation}

The outer loop update is:
\begin{equation}
\label{equation4_17}
\theta\leftarrow\theta-\epsilon\frac1k\sum_{i=1}^k(\theta^{\prime}_i-\theta).
\end{equation}

These optimizations replace the original module with ANIL and Reptile, resulting in the AMA-GAD and ReMA-GAD frameworks, which were tested and compared with the original MA-GAD.

\begin{table}
    \centering
    \caption{Statistical Information of Graph Anomaly Datasets}
    {
    \begin{tabular}{ >{\centering\arraybackslash}m{1.5cm} |
    >{\centering\arraybackslash}m{1cm} 
    >{\centering\arraybackslash}m{1cm} 
    >{\centering\arraybackslash}m{1cm}
    >{\centering\arraybackslash}m{2cm}}
    \toprule
        {Dataset} & \#Graphs & \#Nodes & \#Edges & \#Anomaly Ratio\\ 
        \midrule
        AIDS & 2000 & 31385 & 64780 & 0.2\\
        MUTAG & 188 & 3371 & 7442 & 0.335\\
        PTC-FM & 349 & 4925 & 10110 & 0.41\\
        PTC-MM & 336 & 4695 & 9624 & 0.384\\
        \bottomrule
    \end{tabular}
    }
    \label{table3_2}
\end{table}

\section{EXPERIMENTS}
\subsection{Experimental Setup}
\textbf{Datasets and Metrics.} Four public datasets test the MA-GAD framework, with statistics in Table~\ref{table3_2}. Graphs without attributes use identity matrices as node features. Details are: \textbf{MUTAG~\cite{MUTAG}} (188 compounds) distinguishes carcinogenic (anomalous) from non-carcinogenic compounds; \textbf{AIDS~\cite{AIDS}} classifies HIV-active vs. inactive compounds; \textbf{PTC-FM~\cite{PTC}} and \textbf{PTC-MM~\cite{PTC}} cover carcinogenic vs. non-carcinogenic compounds in female and male mice. The comparison baselines include two types of anomaly detection methods. For graph anomaly detection, baselines are \textbf{GLocalKD~\cite{GLocalKD}}, \textbf{GLADC~\cite{GLAD}}, and \textbf{iGAD~\cite{iGAD}}. For subgraph anomaly detection, baselines include \textbf{GCN-Coarsen\cite{GCN}}, \textbf{HO-GAT\cite{HO-GAT}}, and \textbf{AS-GAE~\cite{AS-GAE}}.

\textbf{Implementation Details.} Graph compression uses a ratio of $r=0.6$, with the GNN anomaly score module having a 256-unit hidden layer. Four auxiliary networks split query and support sets at 50\% each. The target network dataset is split 40\%/20\%/40\% for training/validation/testing. In Equation (\ref{equation8}), $F_{readout}$ is a mean, $z_G=\frac{1}{|V|}\sum_{v \in V_{l}}z_v$. Equation (\ref{equation4_9}) has a 512-unit hidden layer and 1-unit output. In Equation (\ref{equation4_11}), $m=5$. MA-GAD trains for 100 epochs, batch size 8. Meta-training sets $\alpha=0.01$, $\beta=0.008$, with 5 gradient steps for $\theta^{\prime}$. Fine-tuning uses 15 steps. Baselines optimize hyperparameters, and experiments on MA-GAD, AMA-GAD, ReMA-GAD, and baselines repeat 100 times, averaging results.

\begin{table}[t]
    \centering
    \caption{Graph Anomaly Detection Results}
    {
    \begin{tabular}{ >{\centering\arraybackslash}m{1.5cm} |
    >{\centering\arraybackslash}m{1cm} 
    >{\centering\arraybackslash}m{1cm} 
    >{\centering\arraybackslash}m{1.3cm} 
    >{\centering\arraybackslash}m{1.3cm}}
    \toprule
        {Algorithm} & {AIDS} & {MUTAG} & {PTC-FM} & {PTC-MM}\\ 
        \midrule
        iGAD & 0.625 & 0.8000 & 0.4750 & {0.6375}\\
        GlocalKD & 0.3333 & 0.7241 & 0.4833 & 0.5916\\
        GLADC & \textbf{0.9897} & 0.8333 & 0.5666 & 0.4000\\
        \midrule
        CA-GAD & 0.9667 & 0.8413 & 0.5667 & 0.6130\\
        MA-GAD & 0.9817 & 0.9417 & 0.5867 & \textbf{0.6526}\\
        ReMA-GAD & 0.9576 & \textbf{0.9900} & \textbf{0.7086} & 0.5366\\
        AMA-GAD & 0.9833 & 0.9341 & 0.6869 & 0.5832\\
        \bottomrule
    \end{tabular}
    }
    \label{table4_3}
\end{table}

\subsection{Experimental Results on Graph-Level
Anomaly Detection}

\begin{table}[t]
\centering
\caption{Subgraph Anomaly Detection Results}
{
\begin{tabular}{ >{\centering\arraybackslash}m{1.7cm} 
>{\centering\arraybackslash}m{1cm} 
>{\centering\arraybackslash}m{1cm} 
>{\centering\arraybackslash}m{1.3cm} 
>{\centering\arraybackslash}m{1.3cm}}
 \toprule
 {Algorithm} & {AIDS} & {MUTAG} & {PTC-FM} & {PTC-MM}\\ 
 \toprule
GCN-Coarsen & 0.7062 & 0.6642 & {0.7214} & 0.7499\\
HO-GAT & 0.5117 & 0.5015 & 0.4776 & 0.5249\\
AS-GAE & 0.4812 & 0.5350 & 0.5060 & 0.5140\\
 \toprule
CA-GAD & {0.7538} & {0.7844} & {0.7289} & {0.7535}\\
MA-GAD & 0.9028 & 0.9273 & \textbf{0.9746} & 0.8346\\
ReMA-GAD & 0.8623 & \textbf{0.9592} & 0.9231 & \textbf{0.8427}\\
AMA-GAD & \textbf{0.9217} & 0.9133 & 0.9537 & 0.7852\\
\bottomrule
\end{tabular}
}
\label{table4_4}
\end{table}

\textbf{Results of Anomaly Graph Detection} 

The experimental results for anomaly graph detection are presented in Table~\ref{table4_3}, where MA-GAD achieves optimal performance across most datasets, demonstrating its efficacy in detecting anomalous graphs. The analysis is as follows: MA-GAD exhibits high accuracy across all datasets, leveraging meta-learning and graph compression to enhance generalization and embedding quality for superior few-shot anomaly detection. Against baselines, MA-GAD improves ROC-AUC scores by over 0.1 on MUTAG, 0.02 on PTC-FM, and 0.015 on PTC-MM, though it falls short of GLADC on AIDS, still achieving high accuracy. Baseline models GlocalKD and GLADC underperform on some datasets (e.g., GlocalKD’s 0.3 lower score on AIDS), indicating their limitations in unsupervised few-shot scenarios. Comparing MA-GAD with its variants ReMA-GAD and AMA-GAD, all three excel without significant differences, suggesting that ReMA-GAD and AMA-GAD, despite simplified gradient updates, remain effective alternatives.

\textbf{Results of Anomaly Subgraph Detection} 

The MA-GAD framework, compared with baseline methods for anomaly subgraph detection as shown in Table~\ref{table4_4}, achieves superior accuracy across most datasets, with results indicating: MA-GAD exhibits high detection accuracy on all datasets, surpassing the CA-GAD framework, highlighting the efficacy of its meta-learning-based approach; it outperforms baselines by improving ROC-AUC scores by over 0.093, leveraging graph compression and meta-learning for robust performance under limited samples and noise; unsupervised models HO-GAT and AS-GAE yield poor results, nearing random classification in few-shot cases, while supervised GCN-Coarsen enhances embedding quality via graph coarsening but lacks anomaly-specific knowledge and risks overfitting; MA-GAD, ReMA-GAD, and AMA-GAD show competitive performance, with ReMA-GAD and AMA-GAD excelling except on PTC-MM, making them viable alternatives.

\begin{table}[t]
    \centering
    \caption{Ablation Study Results for Graph Anomaly Detection}
    {
    \begin{tabular}{ >{\centering\arraybackslash}m{1.7cm} 
    >{\centering\arraybackslash}m{1cm} 
    >{\centering\arraybackslash}m{1cm} 
    >{\centering\arraybackslash}m{1.3cm} 
    >{\centering\arraybackslash}m{1.3cm}}
    \toprule
        {Algorithm} & {AIDS} & {MUTAG} & {PTC-FM} & {PTC-MM}\\ 
        \midrule
        MA-GAD w/o Meta & 0.9667 & 0.6953 & 0.5667 & 0.5316\\
        MA-GAD w/o Condensation & 0.9376 & 0.9333 & \textbf{0.6237} & 0.5120\\
        MA-GAD & \textbf{0.9817} & \textbf{0.9417} & 0.5867 & \textbf{0.6526}\\
        \bottomrule
    \end{tabular}
    }
    \label{table4_5}
\end{table}

\begin{table}[t]
    \centering
    \caption{Ablation Study Results for Subgraph Anomaly Detection}
    {
    \begin{tabular}{ >{\centering\arraybackslash}m{1.7cm} 
        >{\centering\arraybackslash}m{1cm} 
        >{\centering\arraybackslash}m{1cm} 
        >{\centering\arraybackslash}m{1.3cm} 
        >{\centering\arraybackslash}m{1.3cm}}
    \toprule
        {Algorithm} & {AIDS} & {MUTAG} & {PTC-FM} & {PTC-MM}\\ 
        \midrule
        MA-GAD w/o Meta & 0.8440 & 0.7636 & 0.7093 & 0.7942\\
        MA-GAD w/o Condensation & 0.8553 & 0.8213 & 0.9217 & 0.8282\\
        MA-GAD & \textbf{0.9028} & \textbf{0.9273} & \textbf{0.9746} & \textbf{0.8346}\\
        \bottomrule
    \end{tabular}
    }
    \label{table4_6}
\end{table}
\subsection{Ablation Study}
This subsection examines MA-GAD’s component contributions via ablation studies, comparing the full model against variants with individual components removed. Two variants are tested: \textbf{MA-GAD w/o Meta}, excluding the meta-learning module and training directly on the target graph without auxiliary graph knowledge, and \textbf{MA-GAD w/o Condensation}, omitting graph compression and using GCN on the original graph with a readout module. 

Results for anomaly graph and subgraph detection are in Table~\ref{table4_5} and Table~\ref{table4_6}, showing: (1) MA-GAD outperforms both variants, validating each component’s role; (2) MA-GAD w/o Meta exhibits a larger performance drop than MA-GAD w/o Condensation, highlighting the meta-learning module’s greater significance.
\begin{figure}[t]
    \centering
    
    \begin{minipage}{0.45\textwidth}
    \centering
    \includegraphics[width=\textwidth]{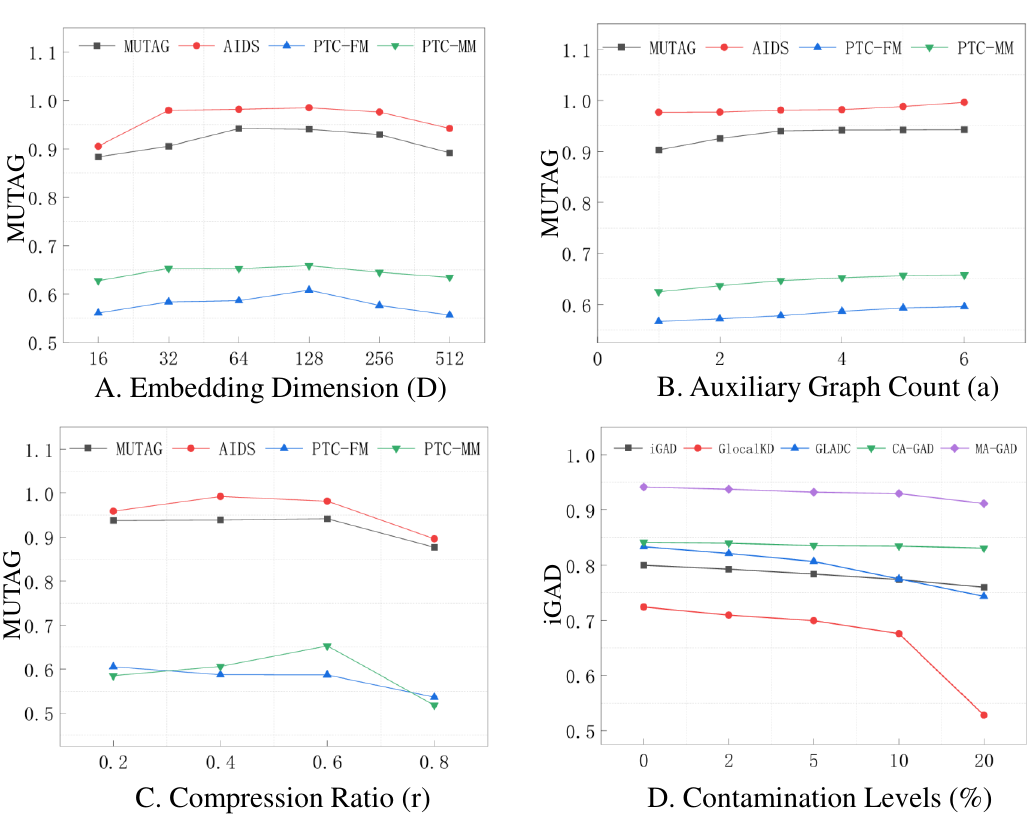}
    \caption{Sensitivity Analysis}
    \label{figure4_3}
    \end{minipage}
\end{figure}

\subsection{Parameter Impact Analysis}

This subsection evaluates the effect of embedding dimension ($D$), auxiliary graph sets ($a$), compression ratio ($r$), and contamination levels on anomaly detection, using anomalous graph detection with multiple parameter settings (see Figure~\ref{figure4_3}). A shows that higher $D$ (e.g., 64 or 128) improves performance, while extreme values degrade it. B indicates larger $a$ enhances ROC-AUC, with gains plateauing as $k$ rises; a moderate $a$ is suggested. C reveals optimal ROC-AUC at $r = 0.4$–0.6, dropping at 0.8 due to information loss. D tests robustness on MUTAG with 0\%–20\% contamination; MA-GAD consistently outperforming others, showing strong robustness.

\subsection{Few-Shot Evaluation}

\begin{table}[t]
    \centering
    \caption{Experimental Results of MA-GAD for Anomaly Graph Detection under $k$-Shot Settings}
    \begin{tabular}{ >{\centering\arraybackslash}m{1cm} |
        >{\centering\arraybackslash}m{1cm} 
        >{\centering\arraybackslash}m{1cm} 
        >{\centering\arraybackslash}m{1.3cm} 
        >{\centering\arraybackslash}m{1.3cm}}
        \toprule
        \textbf{Setting} & \textbf{AIDS} & \textbf{MUTAG} & \textbf{PTC-FM} & \textbf{PTC-MM} \\ 
        \midrule
        1-shot & 0.9673 & 0.9377 & 0.5788 & 0.6471 \\
        2-shot & 0.9817 & 0.9417 & 0.5867 & 0.6526 \\
        4-shot & 0.9887 & 0.9432 & 0.5909 & 0.6545 \\
        8-shot & 0.9910 & 0.9455 & 0.5923 & 0.6632 \\
        \bottomrule
    \end{tabular}
    \label{table4_7}
\end{table}

To assess MA-GAD’s effectiveness in graph anomaly detection under $k$-shot settings (1-shot, 2-shot, 4-shot, 8-shot), we evaluate its performance on the target network using varying numbers of labeled anomalies. Batch sizes are set to 2, 4, 8, and 16 to avoid duplicate labeled anomalies in training batches, with the auxiliary graph configuration unchanged. Table~\ref{table4_7} reports MA-GAD’s AUC-ROC performance across these settings. Results show that even with one labeled anomaly (1-shot), MA-GAD outperforms baselines, with performance improving as labeled anomalies increase, enhancing fine-tuning and detection.

\section{Conclusion}

In this paper, we propose MA-GAD, a meta-learning-based framework for few-shot graph-level anomaly detection, tackling challenges of limited labels, noise, and scarce anomaly knowledge. MA-GAD combines graph compression with meta-learning to reduce graph size while retaining key information and employs auxiliary networks to enhance generalization. Experiments on four biochemical datasets show that MA-GAD outperforms state-of-the-art baselines in both graph and subgraph anomaly detection under few-shot settings, with ablation and sensitivity analyses confirming its robustness and practical effectiveness in $k$-shot scenarios.

\section*{Acknowledgments}
This work was supported by the National Key R\&D Plan of China
(Grant No.2022YFC2602305)

\bibliographystyle{IEEEtran}
\bibliography{reference}

\begin{thebibliography}{10}
\providecommand{\url}[1]{#1}
\csname url@samestyle\endcsname
\providecommand{\newblock}{\relax}
\providecommand{\bibinfo}[2]{#2}
\providecommand{\BIBentrySTDinterwordspacing}{\spaceskip=0pt\relax}
\providecommand{\BIBentryALTinterwordstretchfactor}{4}
\providecommand{\BIBentryALTinterwordspacing}{\spaceskip=\fontdimen2\font plus
\BIBentryALTinterwordstretchfactor\fontdimen3\font minus \fontdimen4\font\relax}
\providecommand{\BIBforeignlanguage}[2]{{%
\expandafter\ifx\csname l@#1\endcsname\relax
\typeout{** WARNING: IEEEtran.bst: No hyphenation pattern has been}%
\typeout{** loaded for the language `#1'. Using the pattern for}%
\typeout{** the default language instead.}%
\else
\language=\csname l@#1\endcsname
\fi
#2}}
\providecommand{\BIBdecl}{\relax}
\BIBdecl

\bibitem{qiao2024deep}
H.~Qiao, H.~Tong, B.~An, I.~King, C.~Aggarwal, and G.~Pang, ``Deep graph anomaly detection: A survey and new perspectives,'' \emph{arXiv preprint arXiv:2409.09957}, 2024.

\bibitem{AS-GAE}
Z.~Zhang and L.~Zhao, ``Unsupervised deep subgraph anomaly detection,'' in \emph{Proceedings of the 22nd IEEE International Conference on Data Mining}.\hskip 1em plus 0.5em minus 0.4em\relax IEEE, 2022, pp. 753--762.

\bibitem{GLocalKD}
R.~Ma, G.~Pang, L.~Chen, and A.~van~den Hengel, ``Deep graph-level anomaly detection by glocal knowledge distillation,'' in \emph{Proceedings of the 15th ACM International Conference on Web Search and Data Mining}, 2022, pp. 704--714.

\bibitem{dou141}
Y.~Dou, K.~Shu, C.~Xia, P.~S. Yu, and L.~Sun, ``User preference-aware fake news detection,'' in \emph{Proceedings of the 44th international ACM SIGIR conference on research and development in information retrieval}, 2021, pp. 2051--2055.

\bibitem{GLAD}
X.~Luo, J.~Wu, J.~Yang, S.~Xue, H.~Peng, C.~Zhou, H.~Chen, Z.~Li, and Q.~Z. Sheng, ``Deep graph level anomaly detection with contrastive learning,'' \emph{Scientific Reports}, vol.~12, no.~1, p. 19867, 2022.

\bibitem{qiu2024self}
C.~Qiu, M.~Kloft, S.~Mandt, and M.~Rudolph, ``Self-supervised anomaly detection with neural transformations,'' \emph{IEEE Transactions on Pattern Analysis and Machine Intelligence}, 2024.

\bibitem{graph2vec144}
A.~Narayanan, M.~Chandramohan, R.~Venkatesan, L.~Chen, Y.~Liu, and S.~Jaiswal, ``graph2vec: Learning distributed representations of graphs,'' \emph{arXiv preprint arXiv:1707.05005}, 2017.

\bibitem{FGSD145}
S.~Verma and Z.-L. Zhang, ``Hunt for the unique, stable, sparse and fast feature learning on graphs,'' \emph{Advances in Neural Information Processing Systems}, vol.~30, no.~1, p.~1, 2017.

\bibitem{DeepFD}
H.~Wang, C.~Zhou, J.~Wu, W.~Dang, X.~Zhu, and J.~Wang, ``Deep structure learning for fraud detection,'' in \emph{Proceedings of the 18th IEEE International Conference on Data Mining}.\hskip 1em plus 0.5em minus 0.4em\relax IEEE, 2018, pp. 567--576.

\bibitem{GCOND}
W.~Jin, L.~Zhao, S.~Zhang, Y.~Liu, J.~Tang, and N.~Shah, ``Graph condensation for graph neural networks,'' \emph{arXiv preprint arXiv:2110.07580}, 2021.

\bibitem{zhao}
B.~Zhao, K.~R. Mopuri, and H.~Bilen, ``Dataset condensation with gradient matching,'' \emph{arXiv preprint arXiv:2006.05929}, 2020.

\bibitem{Deviation_loss}
G.~Pang, C.~Shen, and A.~van~den Hengel, ``Deep anomaly detection with deviation networks,'' in \emph{Proceedings of the 25th ACM SIGKDD International Conference on Knowledge Discovery and Data Mining}, 2019, pp. 353--362.

\bibitem{MAML}
C.~Finn, P.~Abbeel, and S.~Levine, ``Model-agnostic meta-learning for fast adaptation of deep networks,'' in \emph{Proceedings of the 34th International Conference on Machine Learning}.\hskip 1em plus 0.5em minus 0.4em\relax PMLR, 2017, pp. 1126--1135.

\bibitem{ANIL}
A.~Raghu, M.~Raghu, S.~Bengio, and O.~Vinyals, ``Rapid learning or feature reuse? towards understanding the effectiveness of maml,'' \emph{arXiv preprint arXiv:1909.09157}, 2019.

\bibitem{Reptile}
A.~Nichol, J.~Achiam, and J.~Schulman, ``On first-order meta-learning algorithms,'' \emph{arXiv preprint arXiv:1803.02999}, 2018.

\bibitem{MUTAG}
A.~K. Debnath, R.~L. Lopez~de Compadre, G.~Debnath, A.~J. Shusterman, and C.~Hansch, ``Structure-activity relationship of mutagenic aromatic and heteroaromatic nitro compounds. correlation with molecular orbital energies and hydrophobicity,'' \emph{Journal of Medicinal Chemistry}, vol.~34, no.~2, pp. 786--797, 1991.

\bibitem{AIDS}
R.~A. Rossi and N.~K. Ahmed, ``The network data repository with interactive graph analytics and visualization,'' in \emph{Proceedings of the 29th Association for the Advancement of Artificial Intelligence}, 2015.

\bibitem{PTC}
C.~Helma, R.~D. King, S.~Kramer, and A.~Srinivasan, ``{The Predictive Toxicology Challenge 2000-2001},'' \emph{Bioinformatics}, vol.~17, no.~1, pp. 107--108, Jan. 2001.

\bibitem{iGAD}
G.~Zhang, Z.~Yang, J.~Wu, J.~Yang, S.~Xue, H.~Peng, J.~Su, C.~Zhou, Q.~Z. Sheng, L.~Akoglu \emph{et~al.}, ``Dual-discriminative graph neural network for imbalanced graph-level anomaly detection,'' \emph{Advances in Neural Information Processing Systems}, vol.~35, no.~1, pp. 24\,144--24\,157, 2022.

\bibitem{GCN}
T.~N. Kipf and M.~Welling, ``Semi-supervised classification with graph convolutional networks,'' \emph{arXiv preprint arXiv:1609.02907}, 2016.

\bibitem{HO-GAT}
L.~Huang, Y.~Zhu, Y.~Gao, T.~Liu, C.~Chang, C.~Liu, Y.~Tang, and C.-D. Wang, ``Hybrid-order anomaly detection on attributed networks,'' \emph{IEEE Transactions on Knowledge and Data Engineering}, vol.~35, no.~12, pp. 12\,249--12\,263, 2021.

\end{thebibliography}


\end{document}